\documentclass[10pt,twocolumn]{article}

\usepackage{epsfig}
\usepackage{times}
\usepackage{amssymb}
\usepackage{amsmath}
\usepackage{lineno}
\usepackage{hyperref}
\usepackage{multirow}
\usepackage{booktabs}
\usepackage{xcolor}
\usepackage{colortbl}
\usepackage{adjustbox}
\usepackage{graphicx}
\usepackage{subcaption}
\usepackage{pifont}
\usepackage{soul}
\usepackage{microtype}
\usepackage{authblk}
\usepackage[T1]{fontenc}

\hypersetup{
  colorlinks=true,
  linkcolor=blue,
  citecolor=blue,
  urlcolor=blue,
  pdfborder={0 0 0},
}

\title{Context-Aware Network Based on Multi-scale Spatio-temporal Attention for Action Recognition in Videos}
\author[1]{Xiaoyang Li}
\author[1,2]{Wenzhu Yang\thanks{Corresponding author. Email: \texttt{wenzhuyang@hbu.edu.cn}}}
\author[1]{Kanglin Wang}
\author[1]{Tiebiao Wang}
\author[1]{Qingsong Fei}

\affil[1]{School of Cyber Security and Computer, Hebei University, Baoding 071000, Hebei, China}
\affil[2]{Machine Vision Engineering Research Center, Hebei University, Baoding 071000, Hebei, China}

\date{}

\begin{document}
\maketitle

\begin{abstract}
	Action recognition is a critical task in video understanding, requiring the comprehensive capture of spatio-temporal cues across various scales. However, existing methods often overlook the multi-granularity nature of actions. To address this limitation, we introduce the Context-Aware Network (CAN). CAN consists of two core modules: the Multi-scale Temporal Cue Module (MTCM) and the Group Spatial Cue Module (GSCM). MTCM effectively extracts temporal cues at multiple scales, capturing both fast-changing motion details and overall action flow. GSCM, on the other hand, extracts spatial cues at different scales by grouping feature maps and applying specialized extraction methods to each group. Experiments conducted on five benchmark datasets (Something-Something V1 and V2, Diving48, Kinetics-400, and UCF101) demonstrate the effectiveness of CAN. Our approach achieves competitive performance, outperforming most mainstream methods, with accuracies of 50.4\% on Something-Something V1, 63.9\% on Something-Something V2, 88.4\% on Diving48, 74.9\% on Kinetics-400, and 86.9\% on UCF101. These results highlight the importance of capturing multi-scale spatio-temporal cues for robust action recognition.
\end{abstract}

\section{Introduction}\label{sec1}

Action recognition is a crucial task in video understanding, aimed at recognizing human actions within video clips. It has been widely applied in areas such as video surveillance~\cite{khan2024human}, personalized recommendations~\cite{lin2023self}, and human-computer interaction~\cite{yu2024human}. The advancement of deep learning has significantly improved the performance of video action recognition models. Spatial modeling techniques used in image-based tasks can similarly extract spatial information from videos. However, unlike images, videos have an additional temporal dimension, making the capture of spatio-temporal cues essential for effective action recognition.

Actions in the real world often exhibit significant multi-scale characteristics. These actions encompass fine-grained details, such as small shifts and body movements, as well as broader motion patterns. Each scale provides unique information, and effectively capturing and integrating these diverse cues across different scales remains a significant challenge. Most action recognition methods employ two-stream networks~\cite{wang2021multi, luo2021dense} or multipath architectures~\cite{wang2021action, wu2022spatiotemporal, cheng2022cross, liu2019hierarchically} to model spatio-temporal patterns. However, extracting optical flow and utilizing multipath designs significantly increase computational costs. To address this, recent works~\cite{liu2020teinet, wang2021tdn, jiang2023esti} have focused on designing specialized temporal modules for both short-term and long-term motion. Other approaches~\cite{li2020tea, hao2022group, li2025genet} group feature maps along the channel dimension and refine the contextual information within each group to extract spatio-temporal cues from different perspectives. However, these methods often fail to account for the fact that different actions have varying durations and occurring regions, which is crucial for accurate action recognition.

In this paper, we propose a Context-Aware Network (CAN), which introduces an adaptive fusion of multi-scale temporal and multi-scale spatial cues through two complementary modules: the Multi-scale Temporal Cue Module (MTCM) and the Group Spatial Cue Module (GSCM). Unlike previous methods that treat temporal and spatial cues separately, CAN explicitly integrates these cues in a unified framework. The MTCM is designed to capture motion cues at multiple temporal scales, enabling the model to recognize both short-term local movements and long-term action trends. In parallel, the GSCM extracts spatial features at multiple scales by dividing the feature maps into groups, allowing it to capture fine-grained, local, and global spatial patterns. Together, these modules provide a comprehensive representation of motion dynamics, integrating temporal and spatial information across different scales. This coupling ensures adaptive context-awareness by dynamically aligning temporal dynamics with spatial features, providing a more robust spatio-temporal fusion. The key contributions of this work are as follows:
\begin{itemize}
    \item The MTCM was developed as a multi-temporal receptive field structure that efficiently captures temporal cues at different scales, including both fast-changing motion details and overall action flow.
	\item The GSCM was introduced as a grouping structure that extracts spatial cues at multiple scales, thereby enhancing the model's ability to capture fine-grained spatial patterns and contextual information.
	\item We hereby propose the CAN, a unified framework that integrates MTCM and GSCM to capture and fuse multi-scale spatio-temporal cues, achieving competitive performance in action recognition tasks.
\end{itemize}

The remainder of this paper is organized as follows. Section~\ref{sec2} reviews related work. Section~\ref{sec3} provides a comprehensive description of the proposed methodology. Section~\ref{sec4} presents experimental results on five mainstream video action recognition datasets. Finally, Section~\ref{sec5} concludes the paper.

\section{Related work}\label{sec2}

This work is closely related to the areas of spatio-temporal feature learning, attention mechanisms for action recognition, and skeleton-based action recognition. Below, we provide a brief overview of these topics.

\subsection{Spatio-temporal feature learning}\label{subsec2.1}

Spatio-temporal feature learning plays a crucial role in action recognition, as it involves capturing both spatial and temporal dependencies in video data. Early methods, such as 3D CNNs~\cite{duan2020omnisourced, feichtenhofer2020x3d} and dual-stream networks~\cite{wang2021multi, luo2021dense}, model spatio-temporal relations effectively but often struggle to capture long-range dependencies efficiently. Recent advancements~\cite{li2025manet, abdelkawy2025epam} have addressed this challenge by introducing specific modules for spatio-temporal relation modeling. For example, M2A~\cite{gebotys2022m2a} and H-MoRe~\cite{huang2025hmore} focus on improving temporal modeling by leveraging multi-scale temporal structures and human-centric motion representations. These works emphasize the importance of fine-grained temporal modeling, particularly for actions that involve both fast and slow motion patterns. In contrast, our proposed CAN integrates multi-scale temporal and spatial cues through the MTCM and GSCM, enabling a more comprehensive capture of motion cues.

\subsection{Attention mechanisms for action recognition}\label{subsec2.2}

Attention mechanisms~\cite{lehmann2023improving, zhuang2025frequency} have significantly improved action recognition by enabling models to focus on relevant features in both spatial and temporal domains. Models such as ACTION-NET~\cite{wang2021action} and TEA~\cite{li2020tea} utilize temporal attention to capture motion dynamics, while other works like ActNetFormer~\cite{dass2024actnetformer} and MSVL~\cite{chen2024multi} apply global attention across both spatial and temporal domains to enhance model performance. Methods such as GC~\cite{hao2022group} and AGPN~\cite{chen2023agpn} combine multi-attention mechanisms across the channel, temporal, and spatial dimensions to model spatio-temporal patterns effectively. However, many of these methods still face challenges in coordinating multi-scale features. In contrast, our CAN seamlessly integrates spatio-temporal cues at multiple scales, thereby enabling more effective representation learning and improved performance.

\subsection{Skeleton-based action recognition}\label{subsec2.3}

Skeleton-based action recognition~\cite{pan2022view,liu2024decoupled} focuses on modeling human actions using the skeletal structure, which proves particularly useful in low-resolution or noisy video scenarios. Recent advancements have utilized Graph Convolutional Networks (GCNs) to model skeleton data as graphs. For instance, CD-JBF-GCN~\cite{tu2022joint} effectively captures both spatial and temporal dependencies between body joints. Additionally, methods like 2D\(^3\)-SkelAct~\cite{zhang2025robust} combine skeleton features with RGB frames to enhance discriminative ability and improve robustness to occlusions or incomplete pose data. TranSkeleton~\cite{liu2023transkeleton} introduces a hierarchical spatial-temporal transformer, which improves the recognition of complex actions by modeling joint-level features with multi-scale temporal and spatial dependencies. Similarly, the cross-view learning method proposed by~\cite{zheng2021cross} integrates skeleton-based features from multiple viewpoints, improving robustness in multi-view settings and addressing the challenges posed by varying camera angles and poses. Despite these advancements, challenges remain in effectively capturing contextual features across multiple granularities. Our CAN overcomes these limitations by integrating multi-granularity spatio-temporal cues, providing a more robust architecture for action recognition.

\begin{figure*}[t]
	\includegraphics[width=\textwidth]{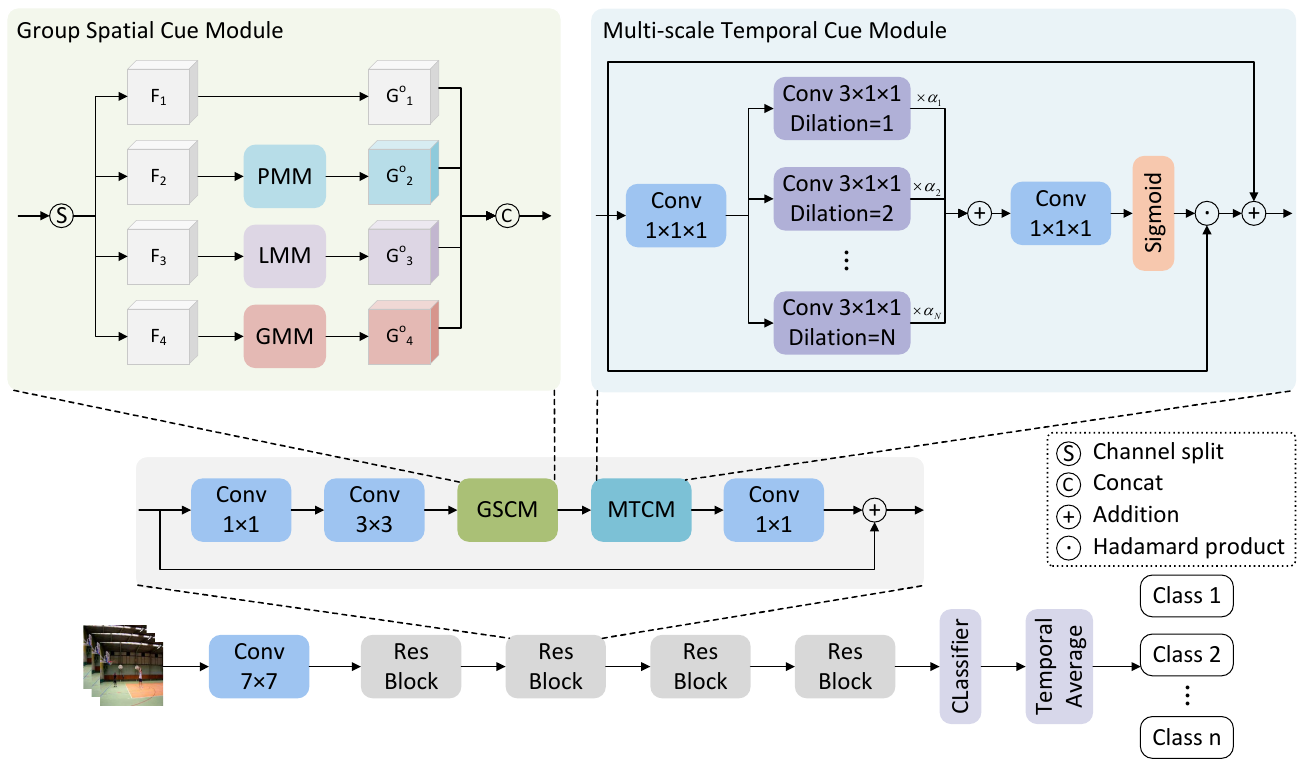}
	\caption{The overall architecture of CAN is instantiated with the ResNet-50 backbone. CAN consists of two key components: the MTCM, which captures motion cues at multiple temporal scales, and the GSCM, which extracts motion cues at various spatial scales.}
	\label{fig1}
\end{figure*}

\section{Methodology}\label{sec3}

This section presents the proposed CAN. First, an overview of CAN is provided, followed by a detailed description of the MTCM and the GSCM.

\subsection{Overview}\label{subsec3.1}

We propose a Context-Aware Network (CAN) for action recognition, designed to effectively capture action cues across multiple spatio-temporal scales. The core idea behind CAN is to enhance the model's ability to recognize actions by leveraging both temporal and spatial context. Specifically, we introduce the Multi-temporal Cue Module (MTCM), which captures rich temporal information by employing a multi-temporal receptive field structure. This structure allows the model to integrate short- and long-term motion cues effectively. In parallel, we design the Group Spatial Cue Module (GSCM) to extract spatial cues at different scales, thereby preserving fine-grained appearance details while incorporating broader contextual information. The combination of MTCM and GSCM allows CAN to capture both dynamic motion patterns and stable spatial features, leading to more accurate and robust action recognition.

First, each video is divided into \( T \) segments, and one frame is randomly sampled from each segment using the sparse sampling strategy~\cite{wang2016temporal}. This results in an input sequence \( \mathbf{I} \) consisting of \( T \) frames, which can be represented as \( \mathbf{I} = [\mathbf{I}_1, \mathbf{I}_2, \dots, \mathbf{I}_T] \). The shape of \( \mathbf{I} \) is \( [T, H, W, C] \), where \( T \) is the number of frames, \( H \) and \( W \) are the height and width of each frame, and \( C \) is the number of feature channels. The feature maps \( \mathbf{F} \) are then extracted from \( \mathbf{I} \) by passing them through the backbone network. Next, \( \mathbf{F} \) is passed through the GSCM and MTCM sequentially within each residual block. The process begins by applying GSCM to capture spatial motion cues at multiple scales. This operation is represented as:
\begin{align}
	\mathbf{G}^{\text{o}} = \mathcal{G}(\mathbf{F})
\end{align}
where \( \mathcal{G} \) denotes the GSCM, and \( \mathbf{G}^{\text{o}} \) is the feature maps obtained from GSCM. Next, the output \( \mathbf{G}^{\text{o}} \) is passed into the MTCM, which extracts motion cues across multiple temporal scales. The operation of MTCM is expressed as:
\begin{align}
	\mathbf{M}^{\text{o}} = \mathcal{M}(\mathbf{G}^{\text{o}})
\end{align}
where \( \mathcal{M} \) represents the MTCM, and \( \mathbf{M}^{\text{o}} \) is the feature maps obtained from MTCM. The final class probability distribution is obtained by a fully connected layer and a temporal consistency function (commonly average pooling) for action classification.

The detailed implementation of both MTCM and GSCM will be provided in the following sections.

\subsection{Multi-scale temporal cue module}\label{subsec3.2}

Temporal cues extracted across multiple temporal scales are essential for action recognition tasks. For instance, accurately recognizing the event ``pretending to close something without actually closing it'' requires not only short-term evidence of the ``closing motion'' but also long-term information about the ``action not actually being completed''. Compared to single-scale temporal modeling methods~\cite{wang2021action, hao2022group}, the MTCM captures both local motion details and overall action trends by utilizing a multi-temporal receptive field design. The key contribution of MTCM lies in its ability to extract motion cues using this multi-temporal receptive field structure.

Fig.~\ref{fig1} illustrates the structure of MTCM. The input feature maps \( \mathbf{G}^{\text{o}} \in \mathbb{R}^{T \times H \times W \times C} \) are first projected into a bottleneck tensor \( \mathbf{M}^{\text{r}} \in \mathbb{R}^{T \times H \times W \times \frac{C}{r}} \) using a \( 1 \times 1 \times 1 \) convolution operation to reduce computational cost. Next, we apply \(N\) parallel depthwise convolutional branches with progressively increasing dilation rates to enhance the model capacity. Specifically, all branches share the same kernel size, while their dilation rates are set to range from 1 to \(N\), enabling multi-scale receptive fields for capturing diverse temporal contexts. To ensure effective collaboration among the multiple branches, each branch output is associated with a learnable weight factor \( \alpha_i \), which controls its contribution to the final fused representation. These weights are initialized uniformly as \( \alpha_i = \frac{1}{N} \), and optimized via backpropagation to adaptively balance the influence of each branch. To stabilize the fusion process and prevent any single branch from dominating, we apply softmax--normalization on the learned weights, ensuring that the sum of the weights remains stable and that each branch contributes proportionally. Specifically:
\begin{align}
	\mathbf{M}^{\text{m}} = \sum_{i=1}^{N} \text{Softmax} (\alpha_i) \times \text{Conv}_{i} (\mathbf{M}^{\text{r}})
\end{align}
where \( \text{Conv}_i \) denotes the convolutional operation for branch \( i \), with dilation rate \( i \), used to capture motion information at a specific temporal scale. Afterward, \( \mathbf{M}^{\text{m}} \) is passed through a \( 1 \times 1 \times 1 \) convolution operation, which projects the channel dimension back to \( C \). The attention weight maps \( \mathbf{M}^{\text{s}} \in \mathbb{R}^{T \times H \times W \times C} \) are then generated by applying a sigmoid activation function. Specifically:
\begin{align}
	\mathbf{M}^{\text{s}} = \sigma \left( \text{Conv}_{1 \times 1 \times 1} (\mathbf{M}^{\text{m}}) \right)
\end{align}
where \( \sigma \) represents the sigmoid activation function. Finally, the attention maps \( \mathbf{M}^{\text{s}} \) are used to emphasize important video frames. To preserve the original features, we apply a residual connection. Specifically:
\begin{align}
	\mathbf{M}^{\text{o}} = \mathbf{M}^{\text{s}} \odot \mathbf{G}^{\text{o}} + \mathbf{G}^{\text{o}}
\end{align}
where \( \odot \) denotes the Hadamard product, and \( \mathbf{M}^{\text{o}} \) is the output feature maps. The resulting attention maps dynamically emphasize informative regions and suppress less useful ones.

In the multi-branch structure of MTCM, the depthwise convolutional branches with different dilation rates extract both long-term action flow features and fine-grained local action changes. The core operation of MTCM is temporal convolution, which allows for the integration of temporal information from adjacent frames. This capability is further enhanced by the multi-temporal receptive field structure, which improves feature representation by integrating motion cues across multiple temporal scales. By learning and normalizing the branch weights, MTCM ensures that the features from different branches collaborate effectively, preserving consistency across time scales while adapting to the most relevant temporal cues for each task.

The branch number \(N\) directly determines the temporal receptive-field coverage of MTCM. Specifically, the \(i\)-th branch adopts a temporal convolution with dilation rate \(i\), whose effective temporal receptive field is \((2^i+1)\). Therefore, increasing \(N\) progressively enlarges the set of temporal scales that can be captured, enabling the model to incorporate longer-range motion cues beyond adjacent frames. However, an excessively large \(N\) may bring diminishing returns: overly long temporal contexts can introduce redundant or noisy dependencies and may weaken the sensitivity to fine-grained short-term motion variations. As validated by the ablation study in Section~\ref{subsubsec4.6.1}, \(N=3\) achieves the best trade-off between temporal coverage and discriminability, providing sufficient modeling capacity for both short- and mid/long-range motion cues without incurring unnecessary computational overhead.

\subsection{Group spatial cue module}\label{subsec3.3}

GSCM is designed to explore motion cues at multiple spatial scales, each serving a specific purpose in capturing spatial features at different levels of granularity. Specifically, we focus on pointwise-level, local-level, and global-level cues. The pointwise-level captures fine-grained spatial details, essential for detecting pointwise-level motion, such as subtle actions like hand gestures. The local-level focuses on slightly larger regions, aggregating local motion patterns that represent more complex actions, while the global-level captures broader spatial context, critical for understanding high-level actions involving large object movements or interactions across a scene.

To preserve the motion patterns learned by the backbone while enriching multi-scale context, we partition the feature maps into four groups along the channel dimension. Specifically, the input feature maps is split into four groups: the first group remains unchanged as an identity path, preserving the original spatial information and acting as a residual anchor that stabilizes fusion and maintains low-level appearance information. The remaining three groups undergo progressively larger receptive-field transformations, enabling hierarchical aggregation of the pointwise-level, local-level, and global-level spatial cues. This design ensures that the GSCM module retains original spatial fidelity while enhancing contextual awareness across multiple granularities, leading to more stable optimization and improved spatial discriminability.

Specifically, let the shape of the input feature maps be denoted as \(\mathbf{F} \in [T, H, W, C]\). First, \(\mathbf{F}\) is divided into four groups, \(\mathbf{F}_i\), where \(i\) denotes the group number. Each group has the shape \([T, H, W, \frac{C}{4}]\). The first group retains the original features, while the remaining three groups apply different spatial cue extraction methods. The operations for each group are as follows:
\begin{align}
    \mathbf{G}_1^o &= \mathbf{F}_1 \nonumber \\
    \mathbf{G}_2^o &= \mathcal{PMM}(\mathbf{F}_2) \nonumber \\
    \mathbf{G}_3^o &= \mathcal{LMM}(\mathbf{F}_3) \nonumber \\
    \mathbf{G}_4^o &= \mathcal{GMM}(\mathbf{F}_4)
\end{align}
where \(\mathbf{G}_i^o \in \mathbb{R}^{T \times H \times W \times \frac{C}{4}}\) is the output feature maps of group \(i\), and \(\mathcal{PMM}\), \(\mathcal{LMM}\), and \(\mathcal{GMM}\) represent the Pointwise Motion Module (PMM), Local Motion Module (LMM), and Global Motion Module (GMM), respectively. Finally, the output feature maps from all groups are concatenated along the channel dimension as follows:
\begin{align}
    \mathbf{G}^o = \text{Concat}(\mathbf{G}_1^o; \mathbf{G}_2^o; \mathbf{G}_3^o; \mathbf{G}_4^o)
\end{align}
where \(\mathbf{G}^o \in \mathbb{R}^{T \times H \times W \times C}\). 

The decision to use four groups in GSCM is motivated by the desired cue granularity and the trade-off between representation capacity and computational cost. Concretely, we design three cue-extraction branches to model motion information at different spatial granularities (fine/pointwise, local, and global), and retain an additional identity branch as a residual anchor to preserve low-level appearance cues and stabilize feature fusion. This functional decomposition naturally leads to a four-path design (three cue-extraction paths plus one identity path), which can be implemented efficiently via channel grouping. Using fewer groups (e.g., three) would force us to remove either the identity path or one cue granularity, thereby weakening cross-scale spatial coverage and feature preservation. Using more groups (e.g., five) would require introducing an additional branch with a clear inductive role, which in practice often entails larger spatial kernels. However, larger kernels (e.g., \(5\times5\) or \(7\times7\)) are less efficient and, as supported by our kernel-size ablation in Section~\ref{subsubsec4.6.2}, do not provide consistent gains under our lightweight design goal. Therefore, we adopt four groups as a balanced and efficient configuration. Finally, concatenating the outputs from all groups forms a rich multi-scale spatial representation that facilitates robust motion recognition across diverse spatial cues.

The detailed operation of each group will be explained in the following sections.

\begin{figure*}[!tb]
	\centering
	\includegraphics[width=0.8\textwidth]{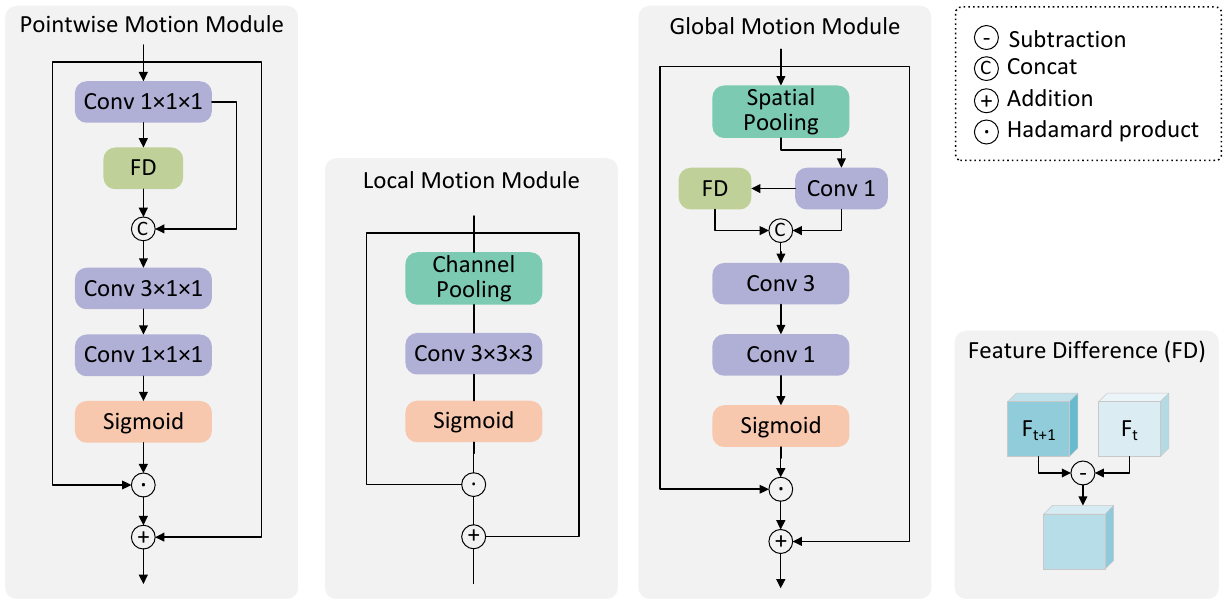}
	\caption{Illustration of each path of GSCM, including Global Motion Module (GMM), Pointwise Motion Module (PMM), and Local Motion Module (LMM).}
	\label{fig2}
\end{figure*}

\subsubsection{Residual connection}

The first group is designed to preserve the original features without applying any additional operations. This group functions as a residual path that maintains the raw spatial information of the input, serving as an identity mapping. This design choice ensures that low-level appearance information from the backbone is preserved and fused with the enhanced spatial cues from the other groups. For actions that do not involve complex spatio-temporal information, such as the ``pick up something'' event, maintaining the original features is sufficient, as no further transformation is necessary. The residual connection stabilizes feature fusion by allowing the network to retain important low-level features, preventing overfitting to the more complex, high-level spatial cues introduced by the other groups. This residual design is key to ensuring that both simple and complex actions can be processed effectively within the same module.

\subsubsection{Pointwise motion module}

The Pointwise Motion Module (PMM), illustrated in Fig.~\ref{fig2}, is employed to capture fine-grained pointwise-level motion cues. The motivation is that subtle motions (e.g., small hand gestures or slight head movements) are often highly discriminative, yet they can be weakened or overlooked when features are aggregated over larger spatial regions. PMM therefore emphasizes motion changes at the smallest spatial scale to better distinguish visually similar actions.

Given the input feature maps \( \mathbf{F}_2 \), we first apply a \(1\times1\times1\) convolution to reduce the channel dimension and obtain
\( \mathbf{G}_2^{\text{r}} \in \mathbb{R}^{T \times H \times W \times \frac{C}{r}} \), which lowers computation while preserving key motion information. To explicitly encode temporal variation, we compute a temporal difference between adjacent frames and concatenate it with the appearance features to form \( \mathbf{G}_2^{\text{c}} \). The concatenated tensor is then processed by a \(3\times1\times1\) convolution to model short-range temporal transitions at each spatial location, followed by a \(1\times1\times1\) projection and a sigmoid function to generate the attention maps \( \mathbf{G}_2^{\text{PMM}} \). Finally, these attention maps recalibrate \( \mathbf{F}_2 \) via a residual formulation:
\begin{align}
    \mathbf{G}_2^\text{r} &= \text{Conv}_{1 \times 1 \times 1}(\mathbf{F}_2) \nonumber \\
    \mathbf{G}_2^\text{c} &= \text{Concat}\left( \mathbf{G}_2^\text{r}; \mathbf{G}_2^\text{r}(t+1) - \mathbf{G}_2^\text{r}(t) \right) \nonumber \\
    \mathbf{G}_2^{\text{PMM}} &= \sigma\left( \text{Conv}_{1 \times 1 \times 1} \left( \text{Conv}_{3 \times 1 \times 1}(\mathbf{G}_2^\text{c}) \right) \right) \nonumber \\
    \mathbf{G}_2^\text{o} &= \mathbf{G}_2^{\text{PMM}} \odot \mathbf{F}_2 + \mathbf{F}_2
\end{align}

Notably, PMM has a spatial receptive field of \(1\times1\), which encourages the module to focus on pointwise-level motion cues. This design is particularly beneficial for actions dominated by minute and rapid changes (e.g., ``picking up an object'' or ``shifting an item slightly''), where fine-grained temporal variations at specific locations are critical for accurate recognition.

\subsubsection{Local motion module}

The Local Motion Module (LMM), shown in Fig.~\ref{fig2}, is used to capture local spatio-temporal motion cues. Unlike the pointwise branch that focuses on per-location variations and the global branch that summarizes the entire frame, LMM targets short-range spatial interactions (e.g., between adjacent joints or nearby regions) that are often critical for distinguishing fine-grained actions. To make local modeling efficient, LMM first removes channel-specific details via channel pooling and then applies a lightweight spatio-temporal convolution on the pooled descriptor.

Given the input feature maps \( \mathbf{F}_3 \), we first pool along the channel dimension to obtain \( \mathbf{G}_3^{\text{pool}} \). We then apply a \(3 \times 3 \times 3\) convolution to extract local spatio-temporal patterns, followed by a sigmoid function to produce the attention maps \( \mathbf{G}_3^{\text{LMM}} \). Finally, the attention maps recalibrate \( \mathbf{F}_3 \) through a residual formulation:
\begin{align}
    \mathbf{G}_3^{\text{pool}} &= \text{ChannelPool}(\mathbf{F}_3) \nonumber \\
    \mathbf{G}_3^{\text{LMM}} &= \sigma\left( \text{Conv}_{3 \times 3 \times 3}(\mathbf{G}_3^{\text{pool}}) \right) \nonumber \\
    \mathbf{G}_3^\text{o} &= \mathbf{G}_3^{\text{LMM}} \odot \mathbf{F}_3 + \mathbf{F}_3
\end{align}

We adopt a \(3 \times 3\) spatial receptive field (within the \(3 \times 3 \times 3\) spatio-temporal kernel) for two reasons. First, \(3 \times 3\) is sufficient to capture short-range motion patterns and local interactions while preserving fine spatial details; in contrast, larger kernels (e.g., \(5 \times 5\) or \(7 \times 7\)) tend to introduce redundant context and blur subtle cues. Second, using \(3 \times 3\) is consistent with the backbone's locality prior, allowing LMM to be integrated without disrupting the receptive-field growth behavior of the network. We further validate this design in our ablation study (Section~\ref{subsubsec4.6.2}), where enlarging the kernel beyond \(3 \times 3\) leads to a decrease in performance, indicating that \(3 \times 3\) offers a better accuracy--efficiency trade-off for local motion modeling.

\subsubsection{Global motion module}

The Global Motion Module (GMM) is applied to capture global motion cues, as illustrated in Fig.~\ref{fig2}. Its goal is to capture \emph{global} motion cues that describe large-scale movements of the human body and interacted objects across the entire scene. Such global dynamics provide complementary evidence to pointwise/local cues and are particularly important for high-level actions with significant spatial displacement (e.g., ``running'' or ``throwing''), where relying only on local changes may be insufficient.

To obtain a compact global motion descriptor, we first squeeze the input feature maps \( \mathbf{F}_4 \) along the spatial dimensions using spatial pooling, resulting in \( \mathbf{G}_4^{\text{pool}} \in \mathbb{R}^{T \times C} \). We then apply a convolution of kernel size \(1\) to reduce channels and obtain \( \mathbf{G}_4^{\text{r}} \in \mathbb{R}^{T \times \frac{C}{r}} \), which improves efficiency while retaining salient motion information. To explicitly encode temporal variation, we compute the temporal difference between adjacent frames of \( \mathbf{G}_4^{\text{r}} \) and concatenate it with the appearance descriptor to form \( \mathbf{G}_4^{\text{c}} \). Next, a 1D temporal convolution with kernel size 3 models short-range temporal transitions, followed by a convolution of kernel size \(1\) back to \(C\) channels and a sigmoid activation to generate attention maps \( \mathbf{G}_4^{\text{GMM}} \). Finally, the attention maps recalibrate the original features via a residual formulation:
\begin{align}
	\mathbf{G}_4^\text{r} &= \text{Conv}_1\left( \text{SpatialPool}\left( \mathbf{F}_4 \right) \right) \nonumber \\
	\mathbf{G}_4^\text{c} &= \text{Concat}\left( \mathbf{G}_4^\text{r} ; \mathbf{G}_4^\text{r}(t+1) - \mathbf{G}_4^\text{r}(t) \right) \nonumber \\
	\mathbf{G}_4^{\text{GMM}} &= \sigma\left( \text{Conv}_1\left( \text{Conv}_3\left( \mathbf{G}_4^\text{c} \right) \right) \right) \nonumber \\
	\mathbf{G}_4^o &= \mathbf{G}_4^{\text{GMM}} \odot \mathbf{F}_4 + \mathbf{F}_4
\end{align}

By operating on the spatially pooled descriptor, GMM emphasizes global motion trends while remaining lightweight. The resulting output \( \mathbf{G}_4^o \) thus encodes global spatial motion cues and provides complementary context for recognizing actions under cluttered backgrounds and large-scale movements.

PMM, LMM, and GMM all operate with a temporal receptive field of size 3, but each module focuses on different spatial scales to capture distinct motion cues. GMM extracts global spatial information by performing spatial squeezing, enabling it to capture large-scale motion patterns across the entire frame. In contrast, PMM emphasizes fine-grained, pointwise-level motion using a spatial receptive field of \(1 \times 1\), which is crucial for recognizing subtle, localized movements. LMM, with its \(3 \times 3\) spatial receptive field, captures local motion cues and spatial details, allowing it to model more intricate spatial interactions. By combining these modules, the model can capture a comprehensive range of motion dynamics across multiple scales: from pointwise-level to local-level and global-level motion cues. This multi-scale integration enhances the representation of motion, enabling the model to effectively recognize actions that involve varying levels of spatial and temporal complexity.

\section{Experiments}\label{sec4}

\subsection{Datasets}\label{subsec4.1}

We conduct experiments on five widely used video action recognition datasets: Something-Something V1~\cite{goyal2017something} and V2~\cite{mahdisoltani2018effectiveness}, Diving48~\cite{li2018resound}, Kinetics-400~\cite{carreira2017quo}, and UCF101~\cite{soomro2012ucf101}. Table~\ref{tab1} summarizes the number of categories and videos for each dataset. Something-Something V1/V2 are temporal-reasoning-dominant benchmarks that emphasize fine-grained object interactions and require modeling subtle temporal cues. Diving48 is a fine-grained dataset with subtle motion differences and strict temporal ordering. Kinetics-400 is a large-scale dataset covering diverse human actions and complex scenes, while UCF101 contains unconstrained real-world videos and is often used to evaluate appearance/scene-biased cases.

Something-Something V1 consists of 108,499 videos, divided into 86,017 for training, 11,522 for validation, and 10,960 for testing, with 174 action categories. Something-Something V2 is a larger and more refined version, containing 220,847 videos, with 168,913 for training, 24,777 for validation, and 27,157 for testing, also with 174 action categories. Diving48 includes 18,404 trimmed clips across 48 categories of competitive diving. Kinetics-400 comprises 400 action categories, with large-scale training data collected from diverse real-world scenarios. UCF101 contains 13,320 videos spanning 101 action categories.

\begin{table*}[!tb]
    \centering
    \footnotesize
    \caption{The number of categories and videos for each dataset, including train/val/test splits.}
    \label{tab1}
    \begin{tabular}{lccccc}
        \hline
        Dataset                & Categories   & Train Videos & Val Videos & Test Videos & Total Videos \\
        \hline
        Something-Something V1 & 174          & 86,017       & 11,522     & 10,960      & 108,499      \\
        Something-Something V2 & 174          & 168,913      & 24,777     & 27,157      & 220,847      \\
        Diving48               & 48           & 14,724       & 1,800      & 2,880       & 18,404       \\
        Kinetics-400           & 400          & 240,000      & 30,000     & 36,245      & 306,245      \\
        UCF101                 & 101          & 9,537        & 1,320      & 2,463       & 13,320       \\
        \hline
    \end{tabular}
\end{table*}

\subsection{Implementation details}\label{subsec4.2}

In our experiments, we use ResNet-50~\cite{he2016deep} as the backbone for implementing CAN, which is pre-trained on ImageNet-1K~\cite{deng2009imagenet}. The factor \( r \) is set to 2 by default. All experiments are performed on an NVIDIA RTX 4070 Ti GPU, with the implementation based on PyTorch 2.0.1 and Python 3.8.

During training, we follow the data augmentation approach in~\cite{wang2016temporal}. Clips are extracted from each video using the sparse sampling strategy, with \( T \in \{8, 16\} \) frames sampled per clip. Each input frame is resized to have its shorter side equal to 256 pixels while maintaining the aspect ratio, followed by a center crop to \( 224 \times 224 \). Optimization is performed using the minibatch stochastic gradient descent (SGD) algorithm with a momentum of 0.9 and weight decay of \( 1 \times 10^{-4} \). The batch size is set to 8, and the training process spans 50 epochs, with the initial learning rate set to 0.0025, decayed by a factor of 10 at epochs 30, 40, and 45. For evaluation, we adopt a 1 clip \( \times \) 1 crop strategy. We sample \( T \in \{8, 16\} \) frames from each video, followed by a \( 224 \times 224 \) center crop. Model performance is reported using Top-1 and Top-5 accuracies.

For 2D CNN-based baselines, we follow the standard evaluation protocol and use the same sampling strategy, data preprocessing, and data augmentation as our implementation to ensure a fair comparison. For some non-2D baselines, such as transformer-based methods, we directly report the results from their original papers, since re-training these large-scale models is computationally prohibitive. To ensure transparency, results quoted from prior work whose experimental settings may differ from ours are explicitly marked with \(^\dagger\). In addition, for methods whose results are not reported in the original papers, we reproduce them under our experimental setup and mark these results with \(^*\).

\begin{table*}[!tb]
	\centering
	\footnotesize
	\caption{Comparison of model performance on Something-Something V1 and V2 datasets. The best results are \textbf{bolded}. ``-'' indicates that the result is not given in the original reference. \(^\dagger\) indicates results quoted from prior work whose training/augmentation or evaluation settings may differ from ours.}
	\begin{tabular}{l c c c c c c c c}
		\hline
		\multirow{2}{*}{Method}              & \multirow{2}{*}{\#Frame} & \multirow{2}{*}{Params (M)} & \multirow{2}{*}{FLOPs (G)} & \multirow{2}{*}{Clips} & \multicolumn{2}{c}{Something-Something V1} & \multicolumn{2}{c}{Something-Something V2} \\
		\cmidrule(lr){6-7} \cmidrule(lr){8-9}
		                                     &                         &                         &                  &        & Top-1 (\%) & Top-5 (\%) & Top-1 (\%) & Top-5 (\%) \\
		\hline
		I3D\(^\dagger\)~\cite{wang2018videos}         & 32             & 28.0                    & 153              & 2      & 41.6  & 72.2 & -    & -    \\
		NL-I3D\(^\dagger\)~\cite{wang2018videos}      & 32             & 35.3                    & 168              & 2      & 44.4  & 76.0 & -    & -    \\
		NL-I3D+GCN\(^\dagger\)~\cite{wang2018videos}  & 32             & 62.2                    & 303              & 2      & 46.1  & 76.8 & -    & -    \\
		S3D-G\(^\dagger\)~\cite{xie2018rethinking}    & 64             & 11.6                    &  71              & 2      & 48.2  & 78.7 & -    & -    \\
		CorrNet-101\(^\dagger\)~\cite{wang2020video}  & 32             & -                       & 224              & 30     & 51.7  & -    & -    & -    \\
		CIDC\(^\dagger\)~\cite{xinyu2020directional}  & 32             & 87.0                    &  92              & 30     & -     & -    & 56.3 & 83.7 \\
		RubiksNet\(^\dagger\)~\cite{fan2020rubiksnet} & 8              & -                       &  33              & 1      & 46.4  & 74.5 & 58.8 & 85.6 \\
		\hline
		TSN~\cite{wang2016temporal}          & 8                       & 23.9                    &  33              & 1      & 19.7  & -    & 33.4 & -    \\
		TSM~\cite{lin2019tsm}                & 8                       & 24.3                    &  33              & 1      & 45.6  & 74.2 & 58.8 & 85.4 \\
		TSM~\cite{lin2019tsm}                & 16                      & 24.3                    &  65              & 1      & 47.3  & 77.1 & 61.2 & 86.9 \\
		SmallBigNet~\cite{li2020smallbignet} & 8                       & -                       &  52              & 1      & 47.0  & 77.1 & 59.7 & 86.7 \\
		SmallBigNet~\cite{li2020smallbignet} & 16                      & -                       & 105              & 1      & 49.3  & 79.5 & 62.3 & 88.5 \\
		TEA~\cite{li2020tea}                 & 8                       & 24.5                    &  35              & 1      & 48.9  & 78.1 & -    & -    \\
		TEA~\cite{li2020tea}                 & 16                      & 24.5                    &  70              & 1      & 51.9  & 80.3 & -    & -    \\
		TDRL~\cite{weng2020temporal}         & 8                       & -                       &  33              & 1      & 49.8  & -    & 62.6 & -    \\
		TDRL~\cite{weng2020temporal}         & 16                      & -                       &  66              & 1      & 50.9  & -    & 63.8 & -    \\
		TANet~\cite{liu2021tam}              & 8                       & 25.1                    &  33              & 1      & 46.5  & 75.8 & 60.5 & 86.2 \\
		TANet~\cite{liu2021tam}              & 16                      & 25.1                    &  66              & 1      & 47.6  & 77.7 & 62.5 & 87.6 \\
		GC-TSN~\cite{hao2022group}           & 8                       & 25.1                    &  33              & 1      & 49.7  & 78.2 & 62.4 & 87.9 \\
		GC-TSN~\cite{hao2022group}           & 16                      & 25.1                    &  67              & 1      & 51.3  & 80.0 & 64.8 & 89.4 \\
		D-TSM~\cite{lee2023d}                & 8                       & -                       &  33              & 1      & -     & -    & 59.8 & 85.9 \\
		ESTI~\cite{jiang2023esti}            & 8                       & -                       &  33              & 1      & 50.2  & \textbf{79.8} & 63.7 & 88.9 \\
		ESTI~\cite{jiang2023esti}            & 16                      & -                       &  66              & 1      & 51.7  & 81.3 & -    & -    \\
		GENet~\cite{li2025genet}             & 8                       & 24.5                    &  33              & 1      & 48.7  & 77.5 & 63.0 & 88.0 \\
		GENet~\cite{li2025genet}             & 16                      & 24.5                    &  67              & 1      & 50.1  & 79.0 & 63.8 & 88.3 \\
		\hline
		CAN (Ours)                           & 8                       & 25.3                    &  35              & 1      & \textbf{50.4} & 78.2 & \textbf{63.9} & \textbf{89.1} \\
		CAN (Ours)                           & 16                      & 25.3                    &  70              & 1      & \textbf{53.9} & \textbf{82.0} & \textbf{64.4} & \textbf{89.7} \\
		\bottomrule
	\end{tabular}
	\label{tab2}
\end{table*}

\begin{table}[!tb]
    \centering
    \footnotesize
    \caption{Comparison of model performance on the Diving48 dataset. The best results are \textbf{bolded}. \(^*\) indicates results reproduced by us (not reported in the original paper). \(^\dagger\) indicates results quoted from prior work whose training/augmentation or evaluation settings may differ from ours.}
	\begin{adjustbox}{max width=\linewidth}
	\begin{tabular}{l l c c}
        \toprule
        Method & Backbone & \#Frame & Top-1 (\%) \\
        \midrule
        TSN from~\cite{hao2022group}                        & ResNet50    & 16    & 78.9 \\
        GST from~\cite{hao2022group}                        & ResNet50    & 16    & 79.8 \\
        TSM from~\cite{hao2022group}                        & ResNet50    & 16    & 83.2 \\
        TDN from~\cite{hao2022group}                        & ResNet50    & 16    & 84.6 \\
        GC-TSN~\cite{hao2022group}                          & ResNet50    & 16    & 86.8 \\
        GC-GST~\cite{hao2022group}                          & ResNet50    & 16    & 82.5 \\
        GC-TSM~\cite{hao2022group}                          & ResNet50    & 16    & 87.2 \\
        GC-TDN~\cite{hao2022group}                          & ResNet50    & 16    & 87.6 \\
        GENet\(^*\)~\cite{li2025genet}                      & ResNet50    & 16    & 82.4 \\
        SlowFast,16\(\times\)8\(^\dagger\) from~\cite{bertasius2021space} & ResNet101   & 64+16 & 77.6 \\
		\midrule
		X3D-M\(^\dagger\) from~\cite{xian2024pmi}           & X3D-M       & 16    & 73.5 \\
		MG-X3D-M\(^\dagger\) from~\cite{xian2024pmi}        & X3D-M       & 16    & 74.6 \\
		PMI-X3D-M\(^\dagger\)~\cite{xian2024pmi}            & X3D-M       & 16    & 81.3 \\
		\midrule
        TimeSformer-HR\(^\dagger\)~\cite{bertasius2021space}& Transformer & 16    & 78.0 \\
        TimeSformer-L\(^\dagger\)~\cite{bertasius2021space} & Transformer & 96    & 81.0 \\
		\midrule
        CAN (Ours)                                          & ResNet50    & 16    & \textbf{88.4} \\
        \bottomrule
    \end{tabular}
	\end{adjustbox}
    \label{tab3}
\end{table}

\begin{table*}[!tb]
    \centering
    \footnotesize
    \caption{Comparison of model performance on the Kinetics-400 dataset. The best results are \textbf{bolded}. \(^*\) indicates results reproduced by us (not reported in the original paper). \(^\dagger\) indicates results quoted from prior work whose training/augmentation or evaluation settings may differ from ours.}
    \begin{tabular}{l c c c c c}
        \toprule
        Method                           & \#Frame & Params (M) & Clips  & Top-1 (\%) & Top-5 (\%) \\
        \midrule
        S3D-G\(^\dagger\)~\cite{xie2018rethinking}   & 64     & 11.6       & 30  & 74.7  & 91.6 \\
        R(2+1)D\(^\dagger\)~\cite{tran2018closer}    & 32     & -          & 10  & 74.3  & 91.4 \\
        I3D\(^\dagger\)~\cite{wang2018videos}       & 64     & 28.0       & -   & 72.1  & 90.3 \\
        ECO\(^\dagger\)~\cite{zolfaghari2018eco}     & 92     & 150.0      & 1   & 70.0  & 89.4 \\
        TSN~\cite{wang2016temporal}      & 25     & 10.7       & 10  & 72.5  & 90.3 \\
        STM~\cite{jiang2019stm}          & 16     & 24.0       & 30  & 73.7  & 91.6 \\
        TSM~\cite{lin2019tsm}            & 16     & 24.3       & 30  & 74.7  & 91.4 \\
        ARTNet~\cite{wang2018appearance} & 16     & -          & 30  & 70.7  & 89.3 \\
        GENet\(^*\)~\cite{li2025genet}   & 16     & 24.5       & 10  & 73.6  & 91.4 \\
        \midrule
        CAN (Ours)                       & 16     & 25.3       & 10  & \textbf{74.9} & \textbf{91.8} \\
        \bottomrule
    \end{tabular}
    \label{tab4}
\end{table*}

\begin{table}[!tb]
	\centering
	\footnotesize
	\caption{Comparison of the mean Top-1 accuracy over three splits on the UCF101 dataset. The best result is \textbf{bolded}.}
	\begin{tabular}{c c}
		\toprule
		Method                                                 & Top-1 (\%)    \\
		\midrule
		TSN~\cite{wang2016temporal}                            & 85.7          \\
		TSM~\cite{lin2019tsm}                                  & 83.2          \\
		TA-VLAD~\cite{sudhakaran2019top}                       & 85.7          \\
		DANet-50~\cite{li2020dual}                             & 86.7          \\
		TDN~\cite{wang2021tdn}                                 & 84.6          \\
		CAN (Ours)                                             & \textbf{86.9} \\
		\bottomrule
	\end{tabular}
	\label{tab5}
\end{table}

\subsection{Comparisons with mainstream methods}\label{subsec4.3}

To evaluate the performance of CAN, we compare it with several mainstream action recognition methods. The experimental results, including accuracy, number of frames (\#Frame), model parameters (Params), and floating-point operations (FLOPs), are presented in Tables~\ref{tab2},~\ref{tab3},~\ref{tab4}, and~\ref{tab5}.

We first compare CAN with baseline networks such as TSN~\cite{wang2016temporal} and TSM~\cite{lin2019tsm}, both of which are based on 2D CNNs. CAN significantly outperforms TSN on the Something-Something V1 and V2, Diving48, Kinetics-400, and UCF101 datasets, primarily due to TSN's limitations in capturing temporal information. TSM improves temporal modeling by shifting channels along the temporal dimension. On Something-Something V1 and V2, CAN improves Top-1 accuracy by 4.8 percentage points (50.4\% vs. 45.6\%) and 5.1 percentage points (63.9\% vs. 58.8\%) with an 8-frame input, and by 6.6 percentage points (53.9\% vs. 47.3\%) and 3.2 percentage points (64.4\% vs. 61.2\%) with a 16-frame input. When using 16-frame input, CAN achieved a Top-1 accuracy improvement of 5.2 percentage points (88.4\% vs. 83.2\%) on the Diving48 dataset and 0.2 percentage points (74.9\% vs. 74.7\%) on the Kinetics-400 dataset. The observed improvements in performance can be attributed to CAN's ability to integrate multi-scale spatio-temporal features. This integration allows the model to capture both short-term motion details and long-term action trends, a key advantage over methods like TSN and TSM that struggle with capturing the full temporal dynamics of actions.

Next, we compare CAN with 2D CNN-based spatio-temporal methods like ESTI~\cite{jiang2023esti} and GENet~\cite{li2025genet}. With an 8-frame input, CAN improves Top-1 accuracy by 0.2 percentage points (50.4\% vs. 50.2\%) and 1.7 percentage points (50.4\% vs. 48.7\%) on Something-Something V1, and by 0.2 percentage points (63.9\% vs. 63.7\%) and 0.9 percentage points (63.9\% vs. 63.0\%) on Something-Something V2. As the number of input frames increases, CAN's performance improves further. For instance, with 16-frame inputs, CAN boosts Top-1 accuracy by 3.8 percentage points (53.9\% vs. 50.1\%) on Something-Something V1 and by 0.6 percentage points (64.4\% vs. 63.8\%) on Something-Something V2, compared to GENet. These results demonstrate CAN's superior ability to capture and model spatio-temporal cues, especially as the input sequence length increases.

We also compare CAN with 3D CNN-based methods~\cite{wang2018videos, wang2020video}. Compared to I3D and S3D-G on Something-Something V1, CAN improves Top-1 accuracy by 8.8 percentage points (50.4\% vs. 41.6\%) and 2.2 percentage points (50.4\% vs. 48.2\%), respectively. On Kinetics-400, CAN surpasses I3D by 2.8 percentage points (74.9\% vs. 72.1\%) in Top-1 accuracy. Additionally, CAN improves Top-1 accuracy by 7.1 percentage points (88.4\% vs. 81.3\%) compared to PMI-X3D-M on Diving48. Finally, when compared to transformer-based methods, such as TimeSformer-L, CAN achieves a 7.4 percentage points (88.4\% vs. 81.0\%) improvement in Top-1 accuracy on Diving48. The significant improvements in CAN's performance across all datasets can be attributed to its ability to effectively model both local and global motion patterns through multi-scale spatio-temporal features. This is particularly evident when compared to methods like TSN and TSM, which have limited capabilities in capturing the full range of temporal dynamics. Furthermore, the robust performance of CAN on more complex datasets, such as Diving48, illustrates its potential to handle a wide variety of action categories with diverse motion characteristics. By leveraging multi-scale features, CAN is able to distinguish between actions with subtle differences, providing a competitive edge over other mainstream methods.

\begin{figure*}[!tb]
    \centering
    \includegraphics[width=0.8\textwidth]{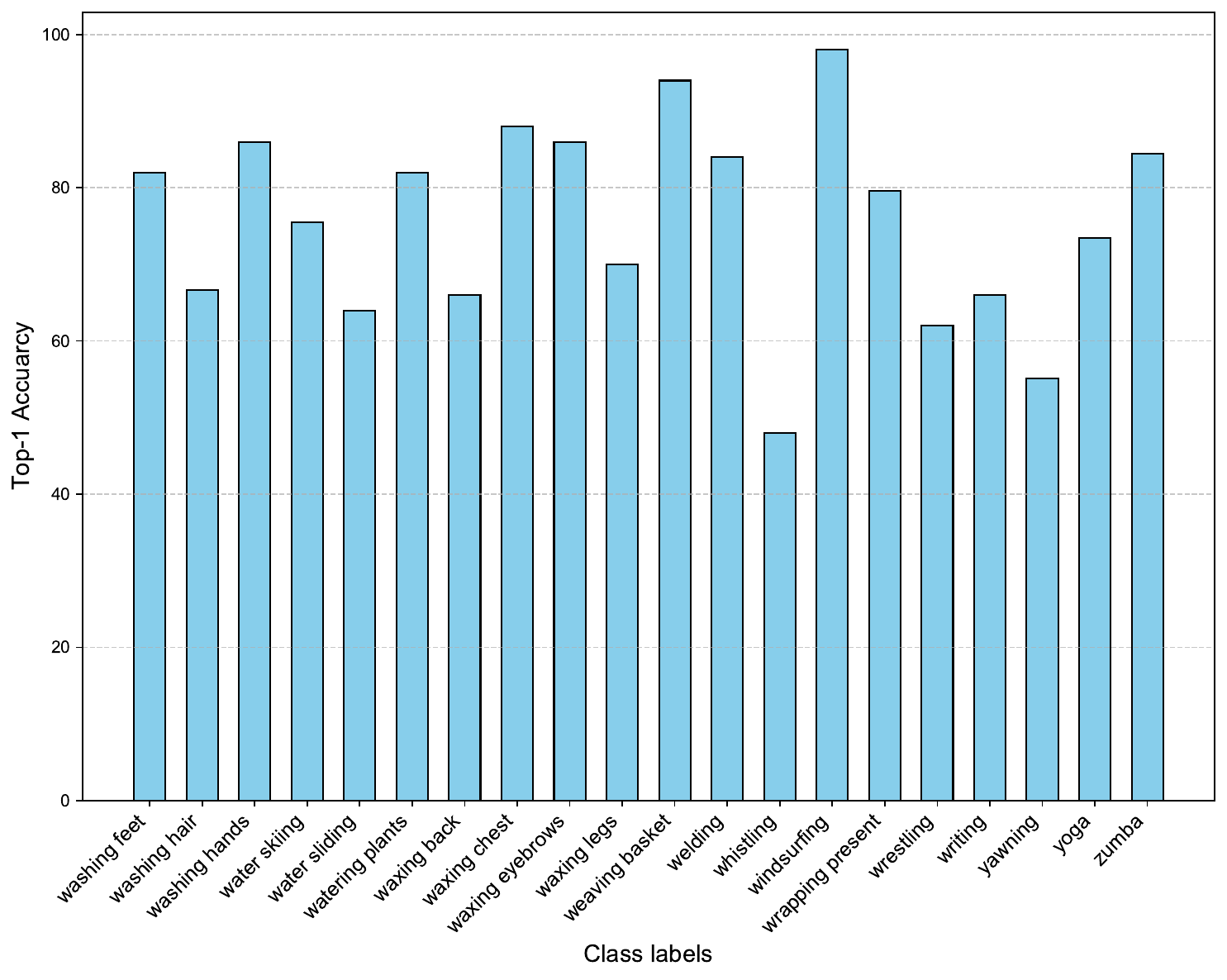}
	\caption{Performance analysis of our CAN on specific action categories from the Kinetics-400 dataset.}
    \label{fig3}
\end{figure*}

\subsection{Performance analysis on specific action categories}\label{subsec4.4}

As shown in Fig.~\ref{fig3}, we further evaluate the generalization ability of CAN on the Kinetics-400 dataset by reporting a per-class performance breakdown. Kinetics-400 is characterized by a long-tail class distribution and cluttered real-world scenes, making it a challenging benchmark for assessing the robustness of spatio-temporal modeling methods.

The per-class results indicate that CAN achieves high accuracy on categories with clear and distinctive motion patterns, such as weaving basket (94.0\%), windsurfing (98.0\%), and washing hands (86.0\%). These actions typically exhibit consistent motion trajectories and well-defined human-object interactions, highlighting the effectiveness of the multi-scale temporal cues captured by MTCM and the spatial cue aggregation enabled by GSCM under complex backgrounds. In contrast, lower accuracy is observed for categories such as whistling (48.0\%) and water sliding (64.0\%), which involve subtle, short-duration motions or ambiguous visual patterns that are easily confused with background context or other actions in cluttered scenes. This performance gap suggests that, while CAN demonstrates strong overall robustness on Kinetics-400, modeling fine-grained or weak motion signals under long-tail distributions remains challenging.

Overall, these results show that CAN maintains competitive Top-1/Top-5 performance on Kinetics-400 while exhibiting consistent strengths across a wide range of action categories. At the same time, the per-class analysis reveals potential limitations in recognizing actions with minimal or ambiguous motion cues. Addressing such cases provides a clear direction for future work, including enhancing temporal sensitivity to subtle motion patterns and incorporating more adaptive mechanisms to better handle long-tail and cluttered scenarios.

\begin{table*}[!tb]
    \centering
    \footnotesize
    \caption{Comparison of inference efficiency for video action recognition models in terms of parameters, FLOPs, latency, and throughput. Less latency and greater throughput represent greater efficiency.}
    \begin{tabular}{l c c c c}
        \toprule
        Method                                   & Params (M) & FLOPs (G) & Latency (ms) & Throughput (video/s) \\
        \midrule
        TimeSformer-HR~\cite{bertasius2021space} & 122.0      & 5110      & 279          & 4  \\
        TimeSformer-L~\cite{bertasius2021space}  & 429.7      & 7140      & 1050         & 1  \\
		\midrule
        TEA~\cite{li2020tea}                     & 24.5       & 35.0      & 25           & 52 \\
        ACTION-Net~\cite{wang2021action}         & 28.1       & 34.8      & 18           & 54 \\
        TDN~\cite{wang2021tdn}                   & 24.2       & 36.2      & 26           & 46 \\
        GC-TDN~\cite{hao2022group}               & 25.4       & 36.8      & 35           & 32 \\
		\midrule
        GSCM (Ours)                              & 24.3       & 33.3      & 24           & 49 \\
        MTCM (Ours)                              & 24.9       & 34.8      & 16           & 80 \\
        CAN (Ours)                               & 25.3       & 35.1      & 32           & 37 \\
        \bottomrule
    \end{tabular}
    \label{tab6}
\end{table*}

\subsection{Runtime performance}\label{subsec4.5}

Table~\ref{tab6} reports a comprehensive comparison of inference efficiency among representative action recognition models, including transformer-based methods, mainstream 2D CNN-based approaches, and our proposed CAN. To provide a realistic assessment of practical efficiency, we evaluate models from multiple perspectives, including parameter count, computational complexity (FLOPs), wall-clock inference latency, and video-level throughput. All latency and throughput measurements are conducted on the same GPU platform (NVIDIA RTX 4070 Ti).

Transformer-based methods, such as TimeSformer-HR and TimeSformer-L, exhibit extremely high computational cost and inference latency. Despite their strong modeling capacity, TimeSformer-HR requires over 5{,}000~G FLOPs per clip and incurs a latency of 279~ms, while TimeSformer-L further increases the computational burden to 7{,}140~G FLOPs with a latency exceeding 1~second, resulting in very low throughput (1--4 videos/s). These results indicate that, in practice, the efficiency of transformer-based architectures is severely limited by memory access patterns and parallelism overhead, making them less suitable for low-latency video inference.

In contrast, CAN demonstrates inference efficiency comparable to mainstream 2D CNN-based methods. Specifically, CAN achieves a latency of 32~ms and a throughput of 37 videos/s, which is on the same order of magnitude as efficient CNN-based approaches such as TEA, ACTION-Net, and TDN. Meanwhile, CAN maintains a relatively low computational cost (35.1~G FLOPs) and a compact parameter size (25.3~M), indicating that the proposed multi-scale spatio-temporal modeling introduces only modest runtime overhead.

These results suggest that CAN effectively balances modeling capacity and computational efficiency. By building upon a 2D CNN backbone and introducing lightweight temporal and spatial cue modeling modules, CAN preserves the favorable latency and throughput characteristics of CNN-based methods, while avoiding the excessive computational and memory overhead associated with transformer-based architectures. This efficiency advantage makes CAN particularly suitable for real-world applications that require low-latency and high-throughput video understanding.

\begin{table}[!tb]
	\centering
	\footnotesize
	\caption{Study on the impact of the number of branches in MTCM using the Diving48 dataset with 8-frame input. The best results are \textbf{bolded}.}
	\begin{adjustbox}{max width=\linewidth}
	\begin{tabular}{l l l c c}
		\toprule
		Configuration    & Params (M)  & FLOPs (G)   & Top-1 (\%)    & Top-5 (\%) \\
		\midrule
		-                & 23.6        & 33.1        & 72.7       & 97.1 \\
		\(N=1\)            & 24.9 (+1.3) & 34.8 (+1.7) & 78.8       & 97.5 \\
		\(N=2\)            & 24.9 (+1.3) & 34.8 (+1.7) & 79.6       & 97.8 \\
		\(N=3\)            & 24.9 (+1.3) & 34.8 (+1.7) & \textbf{81.0} & \textbf{98.2} \\
		\(N=4\)            & 24.9 (+1.3) & 34.9 (+1.8) & 80.1       & 98.0 \\
		\(N=5\)            & 24.9 (+1.3) & 34.9 (+1.8) & 78.5       & 97.8 \\
		\bottomrule
	\end{tabular}
	\end{adjustbox}
	\label{tab7}
\end{table}

\subsection{Ablation studies}\label{subsec4.6}
In this section, we conduct a series of experiments to evaluate the effectiveness of each component of CAN. We train CAN on the Diving48 training set with an 8-frame input and evaluate it on the validation set.

\subsubsection{Study on the impact of the number of branches in MTCM}\label{subsubsec4.6.1}

We investigate the impact of the number of branches \(N\) in MTCM, as shown in Table~\ref{tab7}. The baseline achieves a Top-1 accuracy of 72.7\%. As \(N\) increases, MTCM expands its effective temporal receptive field by introducing branches with larger dilation rates, enabling the model to capture motion cues over longer temporal ranges. Accordingly, performance improves and reaches its peak at \(N=3\), achieving 81.0\% Top-1 accuracy. When \(N>3\), adding more branches brings no further gains. We attribute this to a trade-off between temporal coverage and redundancy: with \(N=3\), MTCM already provides a hierarchical set of receptive fields that cover short-, mid-, and long-range temporal dependencies, which is sufficient for most actions in our benchmarks. Additional branches tend to produce overlapping temporal contexts and introduce extra optimization burden, while contributing limited new discriminative cues. From an efficiency perspective, larger \(N\) also increases computation without commensurate benefits. With \(N=3\), MTCM introduces only 1.3M additional parameters and 1.7G FLOPs, which is modest relative to the baseline. Therefore, considering both accuracy and efficiency, we set \(N=3\) as the default configuration.

\subsubsection{Study on the impact of different spatial receptive field sizes in LMM}\label{subsubsec4.6.2}

In this experiment, we analyze the impact of different spatial receptive field sizes on the performance of the LMM. Specifically, we evaluate kernel sizes of \(3 \times 3\), \(5 \times 5\), and \(7 \times 7\) on the Diving48 dataset with 8-frame inputs, aiming to assess how the receptive field size affects the model's ability to capture local motion cues. As shown in Table~\ref{tab8}, the \(3 \times 3\) kernel yields the best performance, achieving Top-1 accuracy of 79.2\% and Top-5 accuracy of 98.2\%. Increasing the kernel size to \(5 \times 5\) or \(7 \times 7\) results in a performance drop, especially with the \(5 \times 5\) kernel. This is because larger kernels capture excessive context, introducing redundancy and increasing computational cost without significantly improving the model's ability to capture fine-grained motion details. Based on these results, the \(3 \times 3\) kernel is deemed optimal, offering a good balance between performance and computational efficiency.

\begin{table}[!tb]
    \centering
    \footnotesize
    \caption{Study on the impact of different spatial receptive field sizes in LMM using the Diving48 dataset with 8-frame input. The best results are \textbf{bolded}.}
	\begin{adjustbox}{max width=\linewidth}
    \begin{tabular}{l c c c c}
        \toprule
        Kernel Size    & Params (M) & FLOPs (G) & Top-1 (\%)    & Top-5 (\%)    \\
        \midrule	
        \(3 \times 3\) & 23.617      & 33.06    & \textbf{79.2} & \textbf{98.2} \\
        \(5 \times 5\) & 23.618      & 33.06    & 75.7          & 96.8          \\
        \(7 \times 7\) & 23.622      & 33.06    & 79.1          & 97.5          \\
        \bottomrule
    \end{tabular}
	\end{adjustbox}
    \label{tab8}
\end{table}

\begin{table}[!tb]
	\centering
	\footnotesize
	\caption{Study on the effectiveness of each path in GSCM using the Diving48 dataset with 8-frame input. The best results are \textbf{bolded}.}
	\begin{adjustbox}{max width=\linewidth}
	\begin{tabular}{l l l c c}
		\toprule
		Configuration  & Params (M) & FLOPs (G) & Top-1 (\%)    & Top-5 (\%) \\
		\midrule
		-              & 23.61      & 33.05     & 72.7          & 97.1 \\
		PMM            & 23.91      & 33.33     & 78.9          & 98.0 \\
		LMM            & 23.72      & 33.06     & 79.2          & 98.2 \\
		GMM            & 23.91      & 33.06     & 78.8          & 98.0 \\
		\hline
		All paths      & 24.31      & 33.34     & \textbf{80.1} & \textbf{98.2} \\
		\bottomrule
	\end{tabular}
	\end{adjustbox}
	\label{tab9}
\end{table}

\begin{table*}[!tb]
	\centering
	\footnotesize
	\caption{Study on the combination of MTCM and GSCM using the Diving48 dataset with 8-frame input. The best results are \textbf{bolded}.}
	\begin{tabular}{l c c c c}
		\toprule
		Configuration  & Params (M) & FLOPs (G) & Top-1 (\%)    & Top-5 (\%) \\
		\midrule
		-              & 23.6       & 33.1      & 72.7          & 97.1 \\
		GSCM           & 24.3       & 33.3      & 80.1          & 98.2 \\
		MTCM           & 24.9       & 34.8      & 81.0          & 98.2 \\
		Parallelled    & 25.3       & 35.1      & 81.8          & 97.4 \\
		MTCM + GSCM    & 25.3       & 35.1      & 82.1          & 98.1 \\
		GSCM + MTCM    & 25.3       & 35.1      & \textbf{82.7} & \textbf{98.3} \\
		\bottomrule
	\end{tabular}
	\label{tab10}
\end{table*}

\begin{table*}[!tb]
	\centering
	\footnotesize
	\caption{Study on the different embedding configurations using the Diving48 dataset with 8-frame input. The best results are \textbf{bolded}.}
	\begin{tabular}{l l c c}
		\toprule
		Index & Configuration & Top-1 (\%) & Top-5 (\%) \\
		\midrule
		(a) & GSCM \( \to \) MTCM \( \to \) \(1 \times 1\) \( \to \) \(3 \times 3\) \( \to \) \(1 \times 1\) & 78.7 & 97.4 \\
		(b) & \(1 \times 1\) \( \to \) GSCM \( \to \) MTCM \( \to \) \(3 \times 3\) \( \to \) \(1 \times 1\) & 80.5 & 97.6 \\
		(c) & \(1 \times 1\) \( \to \) \(3 \times 3\) \( \to \) GSCM \( \to \) MTCM \( \to \) \(1 \times 1\) & \textbf{82.7} & \textbf{98.3} \\
		(d) & \(1 \times 1\) \( \to \) \(3 \times 3\) \( \to \) \(1 \times 1\) \( \to \) GSCM \( \to \) MTCM & 81.1 & 97.2 \\
		(e) & GSCM \( \to \) \(1 \times 1\) \( \to \) MTCM \( \to \) \(3 \times 3\) \( \to \) \(1 \times 1\) & 80.8 & 98.1 \\
		(f) & GSCM \( \to \) \(1 \times 1\) \( \to \) \(3 \times 3\) \( \to \) MTCM \( \to \) \(1 \times 1\) & 81.4 & 98.2 \\
		(g) & GSCM \( \to \) \(1 \times 1\) \( \to \) \(3 \times 3\) \( \to \) \(1 \times 1\) \( \to \) MTCM & 80.0 & 97.8 \\
		\bottomrule
	\end{tabular}
	\label{tab11}
\end{table*}

\begin{table*}[!tb]
	\centering
	\footnotesize
	\vspace{15pt}
	\caption{Model performance across three backbones using the Diving48 dataset with 8-frame input. TSM is used as the baseline, and CAN demonstrates strong generality across different backbones. The best results are \textbf{bolded}.}
	\begin{tabular}{l l c c c c}
		\toprule
		Backbone & Method     & Params (M) & FLOPs (G) & Top-1 (\%)    & Top-5 (\%)    \\
		\midrule
		\multirow{2}{*}{MobileNet V2}
		         & TSM        & 2.3        & 2.6       & 74.6          & 97.1          \\
		         & CAN (Ours) & 2.5        & 2.8       & \textbf{77.7} & \textbf{97.4} \\
		\hline
		\multirow{2}{*}{ShuffleNet V2}
		         & TSM        & 5.4        & 4.8       & 76.9          & 96.9          \\
		         & CAN (Ours) & 5.8        & 5.6       & \textbf{77.9} & \textbf{97.1} \\
		\hline
		\multirow{2}{*}{ResNet-50}
		         & TSM        & 23.6       & 33.1      & 77.1          & 97.4          \\
		         & CAN (Ours) & 25.3       & 35.1      & \textbf{82.7} & \textbf{98.3} \\
		\bottomrule
	\end{tabular}
	\label{tab12}
\end{table*}

\subsubsection{Study on the effectiveness of each path in GSCM}\label{subsubsec4.6.3}
We evaluate the contribution of each path in GSCM. As reported in Table~\ref{tab9}, introducing any single path consistently improves the baseline, confirming that each component provides useful motion cues for action recognition. Specifically, adding the GMM increases the Top-1 accuracy from 72.7\% to 78.8\%, underscoring the importance of global motion context. Incorporating the PMM yields a comparable gain (72.7\%\(\rightarrow\)78.9\%), suggesting that pointwise-level cues help preserve fine-grained motion variations. The LMM leads to the largest single-path improvement, boosting accuracy to 79.2\%, which we attribute to its effective modeling of local spatio-temporal patterns.

When all paths are enabled jointly, the performance reaches the best result (72.7\%\(\rightarrow\)80.1\%). This indicates that the three motion modules are complementary: global, pointwise-level, and local cues contribute distinct yet synergistic information, and their integration yields a more discriminative spatio-temporal representation.

\subsubsection{Study on the combination of MTCM and GSCM}\label{subsubsec4.6.4}
We investigate how different combinations of MTCM and GSCM affect performance in this section. As shown in Table~\ref{tab10}, introducing either module consistently improves the baseline. Specifically, adding GSCM to the backbone raises the Top-1 accuracy from 72.7\% to 80.1\%, indicating that GSCM provides effective multi-scale spatial cue enhancement. Incorporating MTCM yields an even larger gain, improving the Top-1 accuracy from 72.7\% to 81.0\%, which can be attributed to its multi-temporal receptive-field design for capturing temporal cues at different granularities.

We further compare different fusion strategies of the two modules. The parallel fusion (Parallel) achieves a Top-1 accuracy of 81.8\%. Notably, serial integration delivers higher performance: placing MTCM before GSCM (MTCM\(\rightarrow\)GSCM) achieves 82.1\%, while reversing the order (GSCM\(\rightarrow\)MTCM) reaches the best result of 82.7\%. This suggests that extracting spatial cues first and then refining them with multi-scale temporal modeling leads to more coherent spatio-temporal representations. Therefore, we adopt the GSCM\(\rightarrow\)MTCM configuration as the default setting.

\subsubsection{Study on the different embedding configurations}\label{subsubsec4.6.5}

As shown in Table~\ref{tab11}, MTCM and GSCM can be embedded at seven different locations: (a), (b), (c), (d), (e), (f), and (g). We investigate the impact of each embedding location on the architecture's performance by evaluating the Top-1 and Top-5 accuracy for each configuration. The results reveal that embedding GSCM and MTCM at location (c) yields the highest Top-1 accuracy of 82.7\%, outperforming all other configurations.

Performance drops in locations (a), (b), and (d) highlight the importance of placing GSCM and MTCM after the \(3 \times 3\) convolution. The \(3 \times 3\) convolution is critical for extracting local motion features, which form the foundation for GSCM and MTCM to effectively capture multi-scale motion information. Configurations at locations (e), (f), and (g) show some improvements, but still cannot match the performance achieved by the sequential embedding of GSCM and MTCM at location (c). This suggests that the embedding order significantly impacts performance, as improper sequencing can hinder the effective collaboration between spatial and temporal modules. Based on these findings, we set location (c) as the default embedding configuration, ensuring optimal integration of spatial and temporal features for the best performance.

\begin{figure*}[!tb]
	\centering
	\begin{subfigure}[t]{0.8\textwidth}
		\centering
		\includegraphics[width=\textwidth]{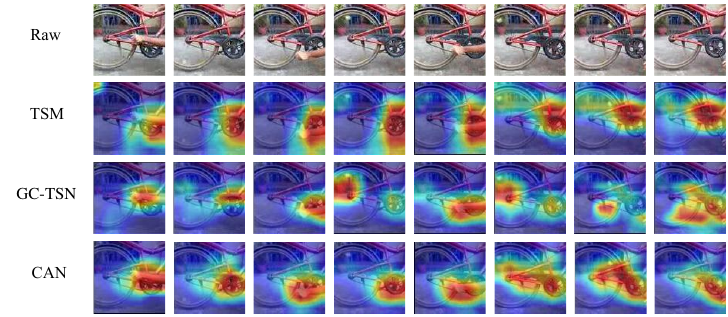}
		\caption{Class: moving part of something}
		\label{fig4a}
		\vspace{5pt}
	\end{subfigure}
	\begin{subfigure}[t]{0.8\textwidth}
		\centering
		\includegraphics[width=\textwidth]{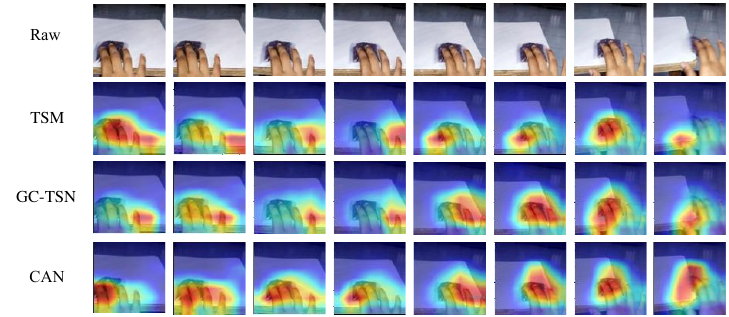}
		\caption{Class: moving something across a surface until it falls down}
		\label{fig4b}
	\end{subfigure}
	\caption{Visualization of important features extracted by TSM, GC-TSN, and CAN. Compared to TSM and GC-TSN, CAN can focus on the region where the action occurs in each frame.}
	\label{fig4}
\end{figure*}

\subsection{Generalization across backbones}\label{subsec4.7}

Similar to TSM~\cite{lin2019tsm}, CAN is designed as a lightweight plug-and-play module that can be seamlessly inserted into standard 2D CNN backbones to enhance spatio-temporal modeling. To verify its flexibility, we evaluate CAN on three representative architectures, including a high-capacity backbone (ResNet-50~\cite{he2016deep}) and two lightweight networks (ShuffleNet V2~\cite{ma2018shufflenet} and MobileNet V2~\cite{sandler2018mobilenetv2}). Following the same setting, we use TSM as the baseline and report model complexity (Params and FLOPs) together with Top-1/Top-5 accuracy.

As shown in Table~\ref{tab12}, CAN consistently outperforms TSM across all three backbones while introducing only a modest increase in computation. In particular, on ResNet-50, CAN yields the largest gain, improving Top-1 accuracy from 77.1\% to 82.7\%. The consistent improvements on both heavy and lightweight backbones demonstrate that CAN is flexible and generalizes well across different network designs.

\subsection{Visualization}\label{subsec4.8}

We employ Grad-CAM~\cite{selvaraju2017grad} to visualize and compare the feature activation maps of TSM, GC-TSN, and our proposed CAN on the Something-Something V1 dataset. Activation maps are generated for all eight input frames to analyze the temporal consistency of model attention. As shown in Fig.~\ref{fig4a}, TSM mainly attends to motion-related regions; however, noticeable misalignments appear in frames 3 and 5, indicating unstable temporal focus. GC-TSN, which introduces a group convolution (GC) module to enhance spatio-temporal modeling, tends to activate scattered and fragmented regions, resulting in less discriminative feature representations. In contrast, CAN consistently concentrates on meaningful hand--object interaction regions, accurately localizing the moving objects across frames. Furthermore, TSM incorrectly emphasizes appearance-dominant regions that are weakly correlated with motion in frames 4 and 8, while GC-TSN shifts its attention to regions that deviate from the true motion areas in frames 3 and 4, as illustrated in Fig.~\ref{fig4b}. By comparison, CAN maintains stable and precise focus on motion-relevant regions throughout all eight frames. These visualizations demonstrate that the multi-scale spatio-temporal cues extracted by CAN enable more coherent and discriminative feature learning, thereby facilitating more reliable action recognition.

\section{Conclusion}\label{sec5}

We propose a simple yet interpretable Context-Aware Network (CAN) for capturing multi-scale spatio-temporal cues in videos. CAN comprises two complementary components: the Multi-scale Temporal Cue Module (MTCM) and the Group Spatial Cue Module (GSCM). MTCM models temporal dynamics across multiple scales via a multi-temporal receptive-field design, enabling it to capture both fast-varying motion details and longer-term action evolution. In parallel, GSCM leverages channel-wise grouping to extract motion cues at different spatial granularities, facilitating the aggregation of fine-to-coarse spatial context. Experiments on five widely used benchmarks demonstrate that CAN consistently outperforms representative mainstream approaches.

\bibliographystyle{main}
\bibliography{main}

\end{document}